\documentclass[journal,twoside,web]{ieeecolor}
\PassOptionsToPackage{unicode}{hyperref}
\PassOptionsToPackage{hyphens}{url}
\usepackage{graphicx}
\usepackage{booktabs}
\usepackage{multirow}
\usepackage{cite}
\usepackage{tabularray}
\usepackage[T1]{fontenc}
\usepackage{graphicx}
\usepackage{times}
\usepackage{epsfig}
\usepackage{graphicx}
\usepackage{amsmath}
\usepackage{amssymb}
\usepackage[mathscr]{eucal}
\usepackage{array}
\usepackage{threeparttable}
\usepackage[ruled]{algorithm2e}

\usepackage{color}
\usepackage{caption}
\usepackage{etoolbox}
\usepackage{generic}
\usepackage{amsmath,amssymb,amsfonts}
\usepackage{textcomp}
\usepackage[pagebackref=true,breaklinks=true,letterpaper=true,colorlinks,bookmarks=false]{hyperref}
\UseTblrLibrary{diagbox}
\usepackage{tmi}
\usepackage{cite}
\usepackage{amsmath,amssymb,amsfonts}
\usepackage{algorithmic}
\usepackage{graphicx}
\usepackage{textcomp}
\def\BibTeX{{\rm B\kern-.05em{\sc i\kern-.025em b}\kern-.08em
    T\kern-.1667em\lower.7ex\hbox{E}\kern-.125emX}}
\begin{document}
\title{Learning to In-paint: Domain Adaptive Shape
Completion for 3D Organ Segmentation}
\author{Mingjin Chen, Yongkang He, Yongyi Lu, Zhijing Yang
}

\maketitle

\begin{abstract}
We aim at incorporating explicit shape information into current 3D organ segmentation models. Different from previous works, we formulate shape learning as an in-painting task, which is named Masked Label Mask Modeling (MLM). Through MLM, learnable mask tokens are fed into transformer blocks to complete the label mask of organ. To transfer MLM shape knowledge to target, we further propose a novel shape-aware self-distillation with both in-painting reconstruction loss and pseudo loss. Extensive experiments on five public organ segmentation datasets show consistent improvements over prior arts with at least 1.2 points gain in the Dice score, demonstrating the effectiveness of our method in challenging unsupervised domain adaptation scenarios including: (1) In-domain organ segmentation; (2) Unseen domain segmentation and (3) Unseen organ segmentation. We hope this work will advance shape analysis and geometric learning in medical imaging.

\end{abstract}

\section{Introduction}
%
%
%
%
\IEEEPARstart{R}{ecently}, Accurately delineating the boundaries of anatomy structures is key to various medical applications such as predicting survival rates and assessing response of tumor microenvironment to various therapeutic techniques such as chemotherapy and radiotherapy. This problem falls into learning the boundary semantics of a structure, i.e., semantic segmentation, or dense classification on the boundary pixels. Segmentation in radiography imaging and photographic imaging is different. For example, nature images are more diverse than radiography images. We can easily distinguish a cat from a dog (e.g., by resorting to the discriminative features such as their appearance differences). However, it is not trivial to distinguish pancreas from other organs given a CT scan since the local contrast between different structures is subtle. 
On the other hand, recent works find that medical image segmentation is more a data problem than a methodology problem~\cite{hofmanninger2020automatic}. Therefore, current sophisticated deep learning segmentation networks proposed in natural images might still fail when directly applied to radiography images. The shape information is beneficial for the segmentation task in medical imaging as different datasets should share the same representation of the 3D anatomy if they are from the same organ (e.g., Pancreas), albeit the change of textures caused by different scanning machines, protocols, phases. Our work seeks to answer this critical question: Can we exploit consistent anatomical shape information to strengthen Deep Nets’ segmentation results, abeit different datasets, image modality and anatomical structures. 

We claim that shape prior is vital for effective segmenting different organs and anatomy structures yet this information is under explored in previous works. One typical example is the prior arts nnU-Net \cite{isensee2021nnu}, which relies heavily on different data augmentations for segmenting different anatomy structures, but none of these operations are shape related. On the other hand, some recent work exploit additional shape priors, e.g., boundary loss \cite{kervadec2019boundary,karimi2019reducing}, signed distance map \cite{xue2020shape}, shape consistency \cite{zhang2022shapepu} and shape template \cite{yang2022implicitatlas} to boost organ segmentation. However, all these methods require the access of label masks in target domain. \cite{yao2022unsupervised} used Variational Autoencoder (VAE) to learn the shape of organs and utilize this to fine-tune segmentation models in an unsupervised way, however, the learned VAE is restricted to remember the mean shape of organ only. 

Unlike existing works, we exploit complementary cue and formulate explicit shape learning as an in-painting task which benefits current Deep Net segmentation models. Specifically, we propose masked label mask modeling (MLM). The label mask of a organ is first divided into visible and corrupted patches and encoded into a series of mask tokens which capture local shape features of organ boundary. Learnable mask tokens are then fed into transformer blocks to complete the label mask of the whole organ. To transfer MLM shape knowledge to target, we propose a novel shape-aware self-distillation with both in-painting reconstruction loss and pseudo loss, though which the organ shape information is transferred from source data to targets in an unsupervised manner. Those target data can be in-domain data (i.e., CT scans), unseen domain data (i.e., MRI) and even segmenting novel class. Extensive experiments on five public organ segmentation datasets show consistent improvements over prior arts.

In summary, We make three primary contributions:
\begin{itemize}
    \item We formulate organ shape learning as an in-painting task via an explicit shape learner called Masked Label Mask Modeling (MLM).
    \item We proposed a shape-aware self-distillation with two new losses to transfer shape knowledge learned by MLM to target.
    \item We demonstrate the effectiveness of our method in challenging unsupervised domain adaptation (UDA) scenarios including: (1) In-domain organ segmentation; (2) Unseen Domain segmentation and (3) Unseen organs segmentation.
\end{itemize}

The rest of this paper is organized as follows. In Section II, we review some related works. The proposed method is introduced in Section III. The experimental results are shown in Section IV. Finally, Section V gives the conclusion.
\begin{figure*}[htbp]
\centering
\includegraphics[scale=0.43]{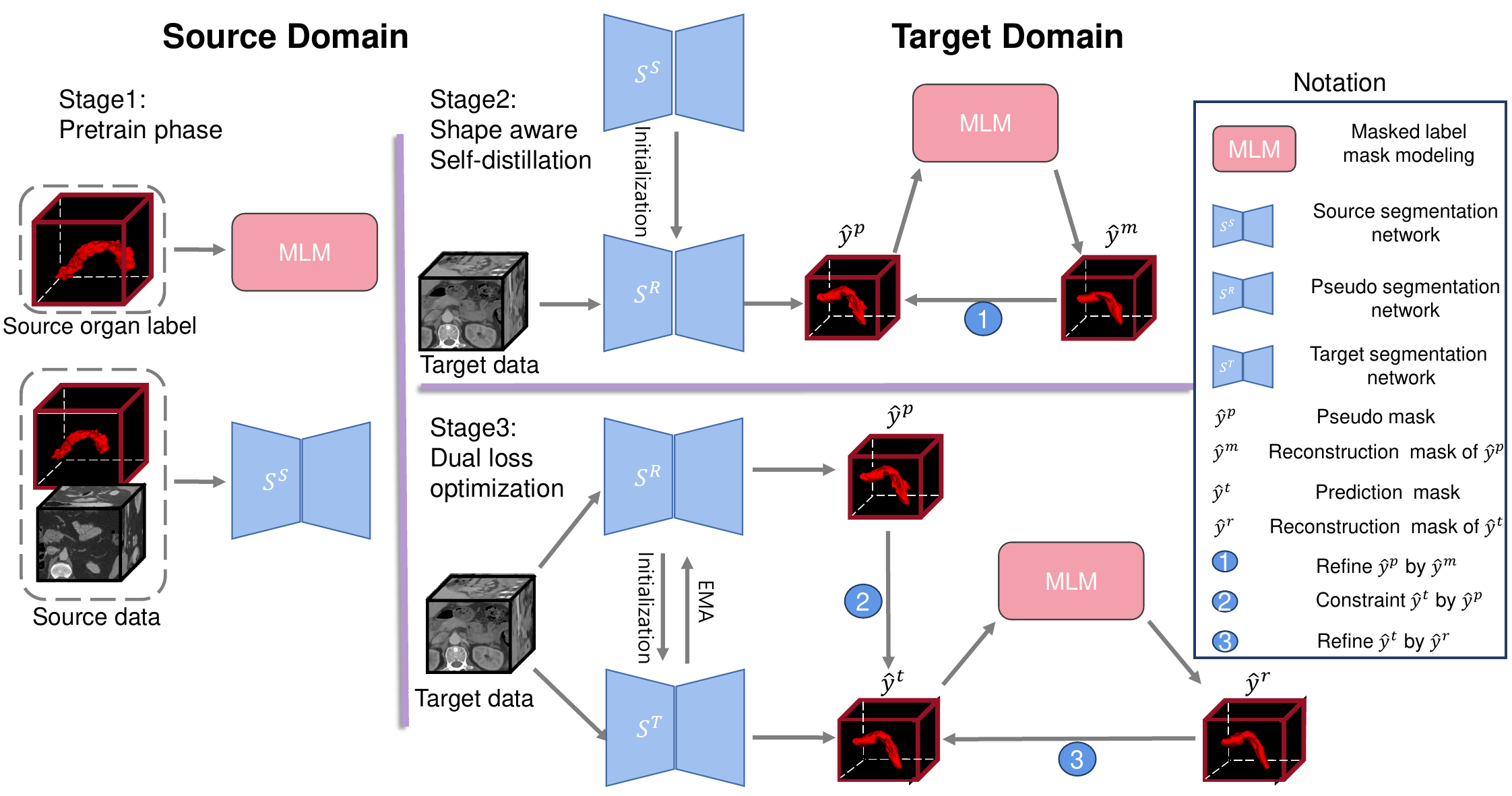}
\caption{Masked label mask modeling pipeline. 
}
\label{proposed MLM-based pipeline}
\end{figure*}

\section{Related Works}

Recently the mainstream of organ segmentation focused on obtaining pre-train model by vision transformer (VIT), or combining the VIT and convolutional neural networks (CNN) more effectively. \cite{zhou2022self,hatamizadeh2022unetr,hatamizadeh2022swin} used the VIT to pretrian model for organ segmentation which achieves a good performance. \cite{xie2021cotr,liu2022phtrans,zhou2021nnformer} combined the strength of CNN and VIT to obtain finer texture. \cite{hofmanninger2020automatic,isensee2021nnu} used various data or data augmentation instead of improving the methodology for organ segmentation.

Unsupervised domain adaptation (UDA) aims to use the domain invariant features of cross-distribution to enhance model generalization in unseen domain. Many works focus on adaption at the image level, image-to-image translation, either converting images from source to target domain \cite{murez2018image,lee2021dranet,bousmalis2017unsupervised,taigman2016unsupervised, chen2020unsupervised} or learning a joint distribution \cite{sankaranarayanan2018generate,kawar2022imagic}, other works focus on adaption at the feature level, which build features that are invariant across domain by minimize the domain gap\cite{long2017deep,wei2019iterative,fu2020domain}. 

In abdominal imaging, the shape of most organs is naturally consistent under various domain shifts. Therefore, some works used organ shape as a reference of cross-distribution to optimize the segmentation results recently \cite{kervadec2019boundary,karimi2019reducing,xue2020shape,yao2022unsupervised,yang2022implicitatlas,zhang2022shapepu}. 
In supervised,  \cite{kervadec2019boundary,karimi2019reducing} have been proposed to constrain the shape boundary of a model's prediction for reducing the difference between the prediction mask and the ground truth mask. 
\cite{xue2020shape} proposed a shape-aware organ segmentation by predicting signed distance maps. ShapePU \cite{zhang2022shapepu} utilizes shape consistency to exploit supervision from unlabeled pixels and capture the global shape features.

In addition, many works focus on modeling the shape  by implicit shape representation \cite{yang2022implicitatlas, raju2022deep, jensen2022deep, khan2022implicit}, which aims to address the label noise or label incomplete problems in the supervised setting. 
In unsupervised, Variational Autocoder (VAE) \cite{liu2019alarm} has been proven that can learn the distribution of shape for a certain organ. Following this work, \cite{yao2022unsupervised} used VAE to learn the shape of organs and utilize this to fine-tune segmentation models in the target domain. However,  previous shape modeling  methods \cite{xue2020shape,yao2022unsupervised,zhang2022shapepu} by contrasting the boundary or learning mean organ shape, which is not useful in large domain gap cases such as different modalities or organs. Unlike all existing works, we are the first work to model shape by an in-painting task.

However,  previous shape modeling  methods \cite{xue2020shape,yao2022unsupervised,zhang2022shapepu} have  focused on the domain gap caused by the cross-datasets. When the domain gap is caused by the unseen domain such as CT to MRI or unseen organs, the existing shape modeling method will have great performance degradation. Our study has proposed a new shape-aware pipeline to transfer shape knowledge to the target model when the domain gap is caused by the unseen domain or unseen organs.

\section{Methods}
\definecolor{g1}{RGB}{65,125,129}
\subsection{Overview}

On unsupervised domain adaptation (UDA) for medical image segmentation, there are typically at least two data sources. The first one, source domain, contains labeled data and is denoted as $\mathcal{D}^{s}=\{x_{i}^{s}, y_{i}^{s}\}_{i=1}^{N}$, where $x_{i}^{s}$  represents an image, $y_{i}^{s}$ represents the corresponding ground truth of image $x_{i}^{s}$, and $N$  is the number of samples in the dataset. The target data, called ``target domain'' does not contain labels and is denoted as $\mathcal{D}^{t}=\{x_{i}^{t}\}_{i=1}^{M}$, where $x_{i}^{t}$ represents an image, and $M$ is the number of samples in the dataset. The aim of UDA is to narrow the domain gap between the source domain  and the target domain.

Previous research has primarily focused on addressing textural variances to narrow the domain gap. However, due to the complexity of texture, it is challenging to obtain a generalized model that performs well on target datasets. Compared to the highly variable texture information in different datasets, the organ shape representation of the 3D anatomy is relatively unchanged. \cite{kervadec2019boundary,karimi2019reducing} constrain the shape by constraining the segmentation boundary, and \cite{yao2022unsupervised} proposed using VAE to learn the distribution of pancreas shapes. 
Different from previous work, we formulate shape learning as an in-painting to exploit the complementary cue. Inspired by the popular masked image modeling (MIM) methods \cite{he2022masked,chen2022context,xie2022simmim,caron2021emerging} we introduce the MLM for modeling shape. In addition to this, shape modeling in the UDA setting usually has two stages, a pre-training stage on the source and an optimization stage in the target domain. However, since the pseudo-labeled model used in the optimization of the target domain is derived from the source domain, this is bound to create noise in the results of the target domain. Therefore in order to reduce the impact of the noise introduced by the models from the source domain, we propose our three-stage framework. The overall architecture of MLM-based pipeline is shown in Figure \ref{proposed MLM-based pipeline}.
\subsection{Masked Label Mask Modeling}
\begin{figure}[htbp]
\centering
\includegraphics[scale=0.38]{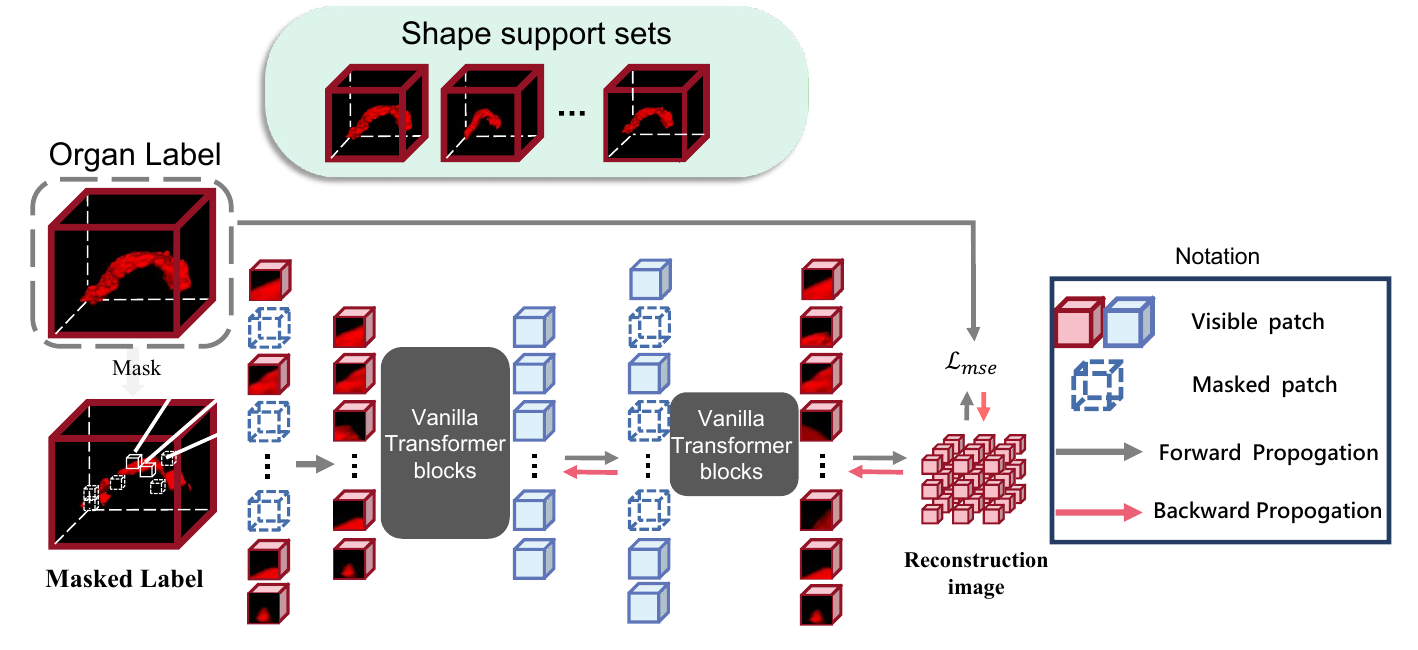}
\caption{Masked Label Mask Modeling. }
\label{MLM}
\end{figure}
Our MLM pipeline consists of three stages. The stage 1 is pretrain phase, we propose a MLM network to learn the shape information with the ground truth mask of the corresponding organ.
And the pretrain segmentation network is also trained on the source domain with CT images and corresponding ground truth mask. In stage 2, we refine the coarse segmentation results obtained from the pseudo segmentation network of the source domain by the results of MLM reconstruction.
This helps to ensure that the pseudo model can generate more accurate pseudo labels by shape information guiding.
In stage 3, we update the target segmentation network by jointly optimizing with MLM reconstruction loss and pseudo loss, in which pseudo loss constrains a shortcut solution of target network and  constrains the segmentation results with low confidence and the recon loss aims to refine the prediction mask.

\noindent\textbf{Pretrained Segmentation Network}
In our propose MLM pipeline, the segmentation network from the source domain serves as initialized model to target domain. The segmentation network can be a pretrained CNN such as 3D-Unet, a vision transformer or other types of segmentator at hand. 
The loss function is Dice score between the segmentation network output $\widehat{y}^{s}$ and the corresponding ground truth $y^{s}$: \begin{equation}
\mathcal{L}_{seg} = - \frac{2\left||{\widehat{y}^{s}\cdot y^{s}}\right||_{1}}{\left||\widehat{y}^{s} \right||_{1} + \left|| y^{s} \right||_{1}}
\end{equation}

\subsubsection{Mask label modeling}
Since most MIM models \cite{he2022masked,chen2022context,xie2022simmim,caron2021emerging,tong2022videomae} have been shown effective in  the pretext task of reconstructing the original image from only partial observations. And, CT volume has a lot of organs, which is not conducive to learning specific organ shapes. Therefore, in order to enable MLM to learn information about the shape of a particular organ, we train the MLM on the ground truth mask $\{y_{i}^{s}\}_{i}^{N}$ in the source domain. Different from most MIM models \cite{he2022masked,chen2022context,xie2022simmim,caron2021emerging,tong2022videomae} which need millions of images to train, our MLM only need a small number of label mark for training.  The MLM module is shown as Fig \ref{MLM}.


Unlike previous mask strategy \cite{he2022masked,chen2022context,xie2022simmim,caron2021emerging,tong2022videomae}  such as random mask, using a large patch, or per-frame masking, we sample the patches which contain the organ pixel which contains the organ local shape feature. Formally, the input organ ground truth mask $G\in \mathbb{R}^{H\times W\times D\times C}$ is first divided into regular non-overlapping 3D patches of $G_{p}\in \mathbb{R}^{N_{p}\times (P^{3}\cdot C)}$, where $C$ is the input channel, $(H, W, D)$ is the organ ground truth mask resolution, and $P$ is the patch size, and $N_{p} = H\cdot W \cdot D / P^{3}$ denote as the numbers of patch. Then we randomly sample the patch which contains the organ pixel to mask with a high mask ratio, the other patches denote as visible patches are fed into the  transformer blocks as encoder $\phi_{enc}$. Finally, learnable mask tokens and patch-wise representations from the encoder are fed into the transformer blocks as decoder $\phi_{dec}$ to obtain the reconstruct  mask $G_r$. The loss function is the mean squared error(MSE) between the pixel-wise input mask and the pixel-wise MLM reconstruction mask:
\begin{equation}
\label{eq:maeloss}
\mathcal{L}_{mse} = \mathsf{\frac{1}{\Omega}\sum_{\Omega}} \left|G - G_{r}\right|^{2}
\end{equation}
where $\mathsf{\Omega}$ is the set of the all patches, $G$ is the input mask, $G_{r}$ is the reconstructed mask.

\subsubsection{Shape-aware Self-distillation}
\begin{figure}[htbp]
\centering
\includegraphics[scale=0.42]{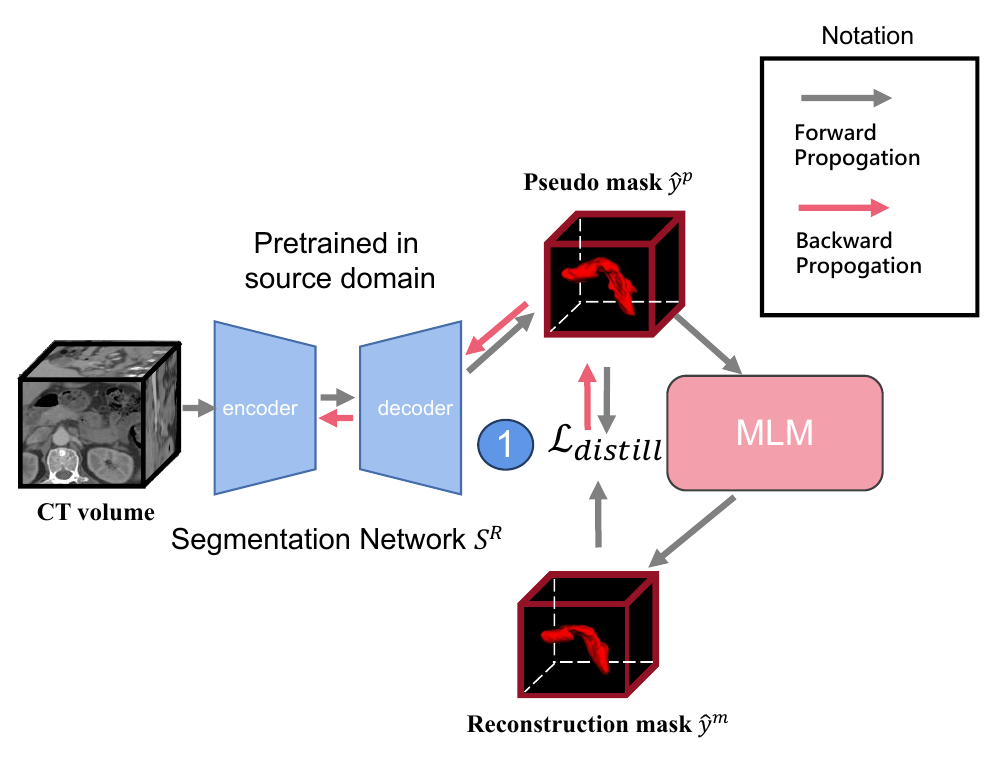}
\caption{Shape-aware self-distillation. }
\label{SA}
\end{figure}
Performing pseudo label by source segmentation model $S^{S}$ for the target domain $\mathcal{D}^{t}$ is one intuitive solution to achieve UDA, which has already been proven effective in semi-supervised learning. However, directly using source segmentation model $S^{S}$ as pseudo model will perform poorly in the target domain, due to the domain gap between the source and target domains.
especially for cases with large domain gaps such as different modalities or different organs.
Therefore, we aim to improve the performance of the segmentation network $S^{S}$ in the target domain by MLM. 


Since our MLM contains the shape information of the target organ, it will not be affected by different textures between different datasets or different modalities, or even different organs. As shown in Fig \ref{SA}, we update the segmentation network $S^S$ by distillation of the shape-aware information from MLM to obtain the segmentation network $S^{R}$ as pseudo model for target model.

Formally the input images $I^{t}$ are fed into the segmentation network $S^{R}$ which Initialization parameters are copied from the source segmentation network $S^{S}$ and obtain the output of the segmentation network $S^{R}$ denote as $\widehat{y}^{p}$. And then, $\widehat{y}^{p}$ as the input is fed into the MLM to obtain the shape-aware output $\widehat{y}^{m}$. We use the Dice coefficient as the loss function between the segmentation network output $\widehat{y}^{p}$ and the MLM output $\widehat{y}^{m}$ denote as $\mathcal{L}_{distill}$. Because the MSE loss focuses on the whole image globally, it represents most of the noise when the organ volume is small. In contrast, the Dice coefficient loss can more accurately measure organ and label differences.:
\begin{equation}
    \label{eq:distill}
    \mathcal{L}_{distill} = - \frac{2\left||{\widehat{y}^{p}\cdot \widehat{y}^{m}}\right||_{1}}{\left||\widehat{y}^{p} \right||_{1} + \left|| \widehat{y}^{m} \right||_{1}}
\end{equation}
the MLM isn't updated in this phase.

\begin{table*}
\centering
\renewcommand\arraystretch{1.4}
\refstepcounter{table}
\captionsetup{}
\caption{Performance comparison of MLM pipeline with other UDA segmentation methods using the different backbones. The segmentation results are evaluated with mean Dice score. $\uparrow$ indicates improvement compared to the best comparison method.}
\label{t1}
\begin{tabular}{c|cc|cc|cc} 
\hline
\multicolumn{1}{l|}{\diagbox{{[}method]}{{[}dataset]}} & \multicolumn{2}{c|}{~MSD}                                                      & \multicolumn{2}{c|}{synapse}                                                   & \multicolumn{2}{c}{WORD}                                                                 \\ 
\hline
backbone                                                & \textbf{3D-Unet}                  & \textbf{Swinunter}                         & \textbf{3D-Unet}                  & \textbf{Swinunter}                         & \textbf{3D-Unet}                           & \textbf{Swinunter}                          \\
Direct test~~                                           & 0.6788                            & 0.6783                                     & 0.7750                            & 0.7198                                     & 0.6809                                     & 0.7473                                      \\
Boundary  \cite{kervadec2019boundary}                                                & 0.6837                            & 0.6628                                     & 0.7794                            & 0.7092                                     & 0.6993                                     & 0.7419                                      \\
Hausdorff distance \cite{karimi2019reducing}                                     & 0.6981                            & 0.6704                                     & 0.7863                            & 0.7218                                     & 0.7226                                     & 0.7220                                      \\
VAE \cite{yao2022unsupervised}                                                    & 0.7502                            & 0.7289                                     & 0.7867                            & 0.7729                                     & 0.7463                                     & 0.7621                                      \\
DISSM \cite{raju2022deep}                                                  & 0.7372                            & 0.7109                                          & 0.7895                            & 0.7682                                          & 0.7480                                     & 0.7648                                           \\
SIFA \cite{chen2020unsupervised}                                                   & 0.6608                            & -                                          & 0.7563                            & -                                          & 0.7280                                     & -                                           \\

nnU-Net   \cite{isensee2021nnu}                                              & 0.6405                            & -                                          & 0.7706                            & -                                          & 0.7691                                     & -                                           \\
MLM(ours)                                               & \textbf{0.7825}($\uparrow$0.0323) & \textbf{\textbf{0.7708}}($\uparrow$0.0419) & \textbf{0.8020}($\uparrow$0.0125) & \textbf{\textbf{0.7848}}($\uparrow$0.0119) & \textbf{\textbf{0.7946}}($\uparrow$0.0254) & \textbf{\textbf{0.7955}}($\uparrow$0.0333)  \\
upper bound                                             & 0.8246                            & 0.8320                                     & 0.8037                            & 0.7975                                     & 0.8403                                     & 0.8484                                      \\
\hline
\end{tabular}
\end{table*}

\subsubsection{Dual-loss Optimization}
\begin{figure}[htbp]
\centering
\includegraphics[scale=0.34]{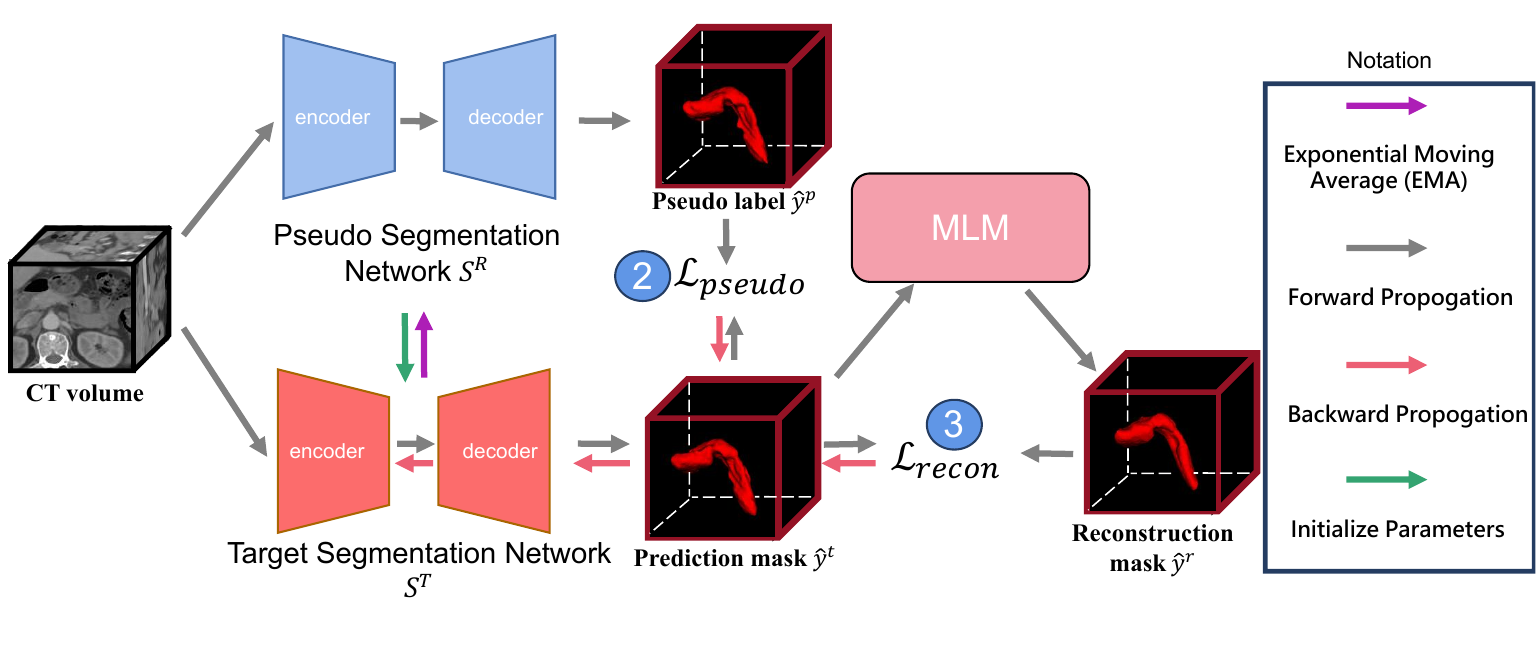}
\caption{Dual-loss Optimization. }
\label{DLO}
\end{figure}
As shown in Fig \ref{DLO}, we use the segmentation network $S^{R}$ as the pseudo model for target model. 
To obtain the target segmentation network $S^{T}$, the parameters of the pseudo model are copied to the target model as the initialization parameters. In order to obtain the target model $S^{T}$ with the explicit model of shape, the pipeline includes the MLM, aiming to introduce the organ shape prior to reduce the biases with the $\mathcal{L}_{recon}$:
\begin{equation}
    \label{eq:reconloss}
    \mathcal{L}_{recon} = - \frac{2\left||{\widehat{y}^{t}\cdot \widehat{y}^{r}}\right||_{1}}{\left||\widehat{y}^{t} \right||_{1} + \left|| \widehat{y}^{r} \right||_{1}}
\end{equation}

However, the loss $\mathcal{L}_{recon}$ is only related to the MLM, which might lead to the target model collapse. Therefore, another loss is required to prevent the segmentation network $S^{T}$ collapse. 

The pseudo label which is the output of the pseudo model $S^{R}$ can serve as the constraint to the target model. The loss function $\mathcal{L}_{pseudo}$ is Dice loss between the pseudo label and the target model output $\widehat{y}^{t}$:
\begin{equation}
    \label{eq:tpseudoloss}
    \mathcal{L}_{pseudo} = - \frac{2\left||{\widehat{y}^{t}\cdot \widehat{y}^{p}}\right||_{1}}{\left||\widehat{y}^{t} \right||_{1} + \left|| \widehat{y}^{p} \right||_{1}}
\end{equation}
The impact of the pseudo-model $S^{R}$ is in two aspects. First, the pseudo-model can provide a more accurate pseudo-label and prevent target model to generate an identical predicted output that has small reconstruction loss. Second, the combined effect of the pseudo label and the reconstructed image enables the target model to learn the shape information of the organ more accurately.
Based on the reconstruction loss and the pseudo loss, the target model loss can be defined as:
\begin{equation}
\label{eq:totalloss}
    \mathcal{L}_{\theta} = \lambda_{pseudo} \cdot \mathcal{L}_{pseudo}  + \mathcal{L}_{recon}
\end{equation}
where $\lambda_{pseudo}$ is a hyperparameter mediating the two losses. $\mathcal{L}_{recon} $ is the reconstruction loss which computes between the target model output $\widehat{y}^{t}$ and the MLM output $\widehat{y}^{r}$. After, the pseudo model $S^{R}$ update by the exponential moving average, the update rule is $ \theta_{r} \leftarrow \beta\theta_{r} + (1-\beta)\theta_{t} $, where $\beta \in[0,1) $  is a momentum coefficient, $\theta_{r}$ is the parameters of the pseudo model $S^{R}$, $\theta_{t}$ is the parameters of the target model $S^{T}$.  
\begin{algorithm}[t]  
	\caption{Our proposed MLM for another organ}
 \label{algorithm}
	\LinesNumbered 
	\KwIn{one sample of another organ data $(G_{s})$, unlabeled target data $I_{t}$, total iterations $N$, EMA update iteration $M$, preatrain MLM, source segmentation network $S^{S}$.  \\}
	\KwOut{Target segmentation model $S^{T}$}
  Fine-tune the MLM on another organ data with $\mathcal{L}_{mse}$. \\
  
  Obtain $S^{R}$ by fine-tuning the source segmentation network $S^{S}$ on unlabeled target data with fine-tune MLM and $\mathcal{L}_{distill}$. \\
  Initialize parameters $S^{T}$ as well as $S^{R}$. \\
    \For{$n=1 \text{ to } N$}{
     Randomly select a batch of target data $I_{t}$. \\
     compute the pseudo loss $\mathcal{L}_{pseudo}$
     between pseudo label and prediction.\\
     compute the recon loss $\mathcal{L}_{recon}$
     between reconstruction of prediction and prediction.\\
     update the model $S^{T}$ by computing the total loss. \\
    \If{n \% M == 0}
      {
        update the model $S^{R}$ by EMA rule. \\
  }
     
	}

\end{algorithm}

\section{Experiments}


We use five public pancreas CT datasets(NIH, MSD, Synapse, AMOS, WORD), and use the NIH as the source domain and pancreas as the source organ for pertaining the MLM and pretrain segmentation model, the other four datasets serve as the target domain data for testing. In Sec.\ref{dataset} we provide more information on the datasets used in our experiments. In Sec.\ref{implementation}, we provide the implementation details in our experiments. In Sec.\ref{e1}, we first validate the MLM in the in-domain pancreas segmentation with three target datasets(MSD, Synapse, WORD) and perform the ablation experiments on MSD. In Sec.\ref{e2}, Sec.\ref{e3} we validate the performance of MLM in unseen domain on AMOS and other unsenn organs on WORD. 
\subsection{Datasets and Data Preprocessing}
\label{dataset}
\textbf{NIH Datasets \footnote{\href{https://wiki.cancerimagingarchive.net/display/Public/Pancreas-CT}{https://wiki.cancerimagingarchive.net/display/Public/Pancreas-CT}}}
Pancreas CT contains 82 abdominal contrast enhanced 3D CT scans. The CT
scans have resolutions of 512 × 512 pixels with varying pixel sizes and slice thickness
between 1.5 $\sim$  2.5mm, acquired on Philips and Siemens MDCT scanners. The dataset
is randomly splitted into a training set of 61 training cases and 21 test cases.

\textbf{MSD Datasets \footnote{\href{http://medicaldecathlon.com/}{http://medicaldecathlon.com/}}}
\cite{antonelli2022medical} contains 420 portal-venous phase 3D CT scans (282 Training and 139 Testing),
having labels of pancreas and tumor. The CT scans have resolutions of 512 × 512 × $l$
pixels. We merge the pancreas and tumor labels together as pancreas in our task. As
we do not know the annotation on the test data, we randomly split the training set
into our training set of 210 cases and test set of 72 cases.

\textbf{Synapse Datasets \footnote{\href{https://www.synapse.org/\#!Synapse:syn3193805/wiki/217789}{https://www.synapse.org/\#!Synapse:syn3193805/wiki/217789}}}
Synapse contains 50 abdomen CT scans (30 Training and 20 Testing). Each CT
volume consists of 85 $\sim$ 198 slices of 512 × 512 pixels, with a voxel spatial resolution
of ([0.54 $\sim$ 0.54]×[0.98 $\sim$ 0.98]×[2.5 $\sim$ 5.0])mm$^3$
. As we do not know the annotation
on the test data, we randomly split the training set into our training set of 22 cases
and test set of 8 cases.

\textbf{AMOS Datasets\footnote{\href{https://amos22.grand-challenge.org/Dataset/}{https://amos22.grand-challenge.org/Dataset/}}}
\cite{ji2022amos} contains  total of 600 CT and MRI( 200 CT and 40 MRI training cases,  100 CT + 20 MRI validation cases, and 200 CT + 40 MRI testing cases.) . The CT and MRI have different resolutions with varying pixel sizes and with a voxel spatial resolution of  ([0.45 $\sim$ 1.06]×[0.45 $\sim$ 1.06]×[1.25 $\sim$ 5.0])mm$^3$, and acquired on different scanners such as Aquilion ONE, Optima CT660, Signa HDe and so on.  As we do not know the annotation
on the test and validation data, we randomly split the CT training set into our training set of 159 cases
and testing set of 41 cases, all the MRI cases belong to the test set.

\textbf{WORD Datasets\footnote{\href{https://github.com/HiLab-git/WORD}{https://github.com/HiLab-git/WORD}}}
\cite{luo2022word} contains 150 abdomen CT scans (100 Training, 20 validation, and 30 Testing). 
Each CT volume consists of 159 to 330 slices of 512 × 512 pixels, with an in-plane
resolution of 0.976 mm × 0.976 mm and slice spacing of 2.5 mm to 3.0 mm, acquired on a Siemens CT scanner. As we do not
know the annotation on the test data, we use the 20 cases of the validation of WORD as our test set.
\begin{figure*}
\centering
\includegraphics[scale=0.38]{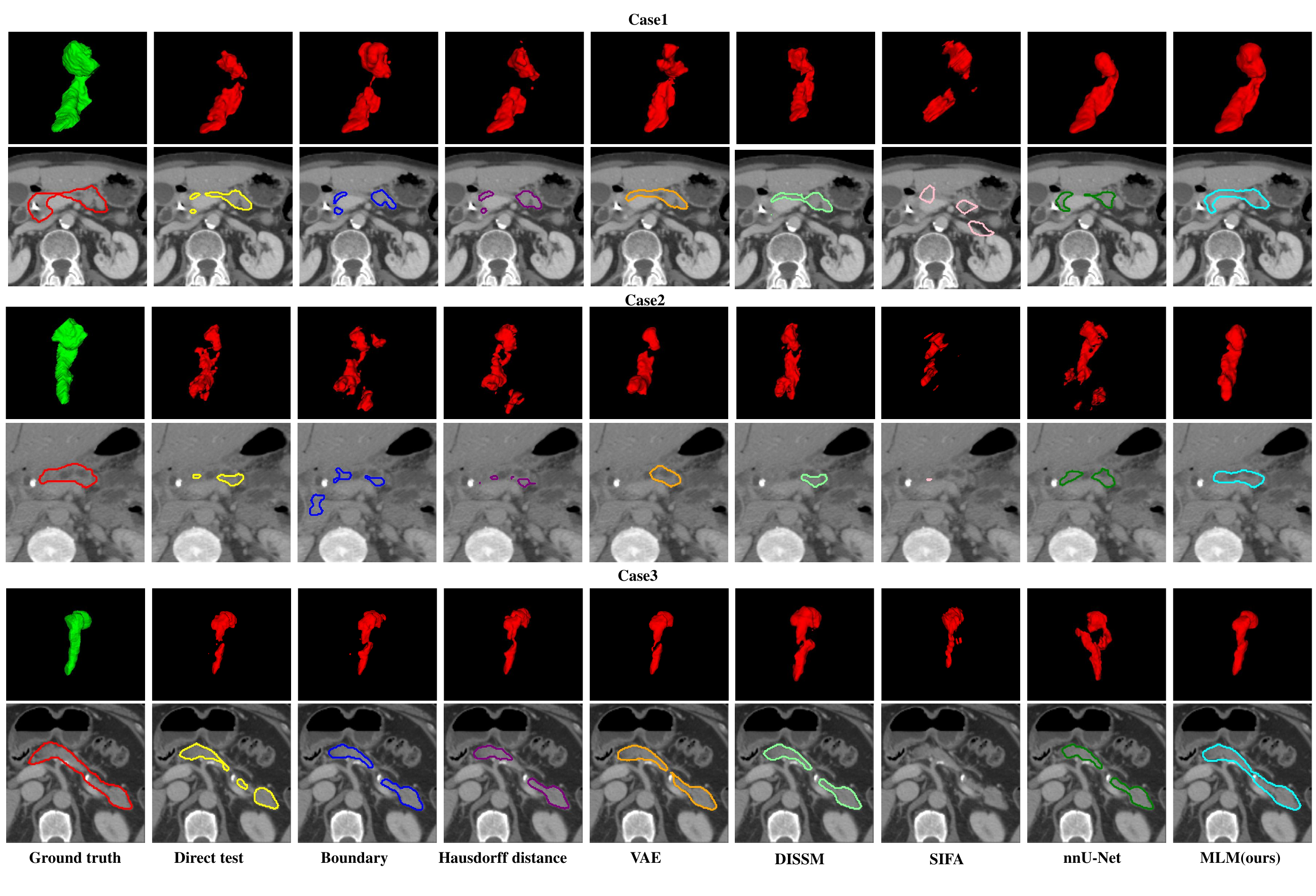} 
\caption{The 3D corresponding 2D visualization for the results on  MSD test cases. For each case, the top image is a 3D visualization, and the bottom image is a corresponding 2D visualization. 
}
\label{result}
\end{figure*}
\textbf{Data Preprocessing}
We convert the intensities to a range of -200 to 400, then further normalize
them into -1 to 1. Image size of 128 × 128 × 128 is applied to all training data. We apply
augmentations including random intensity scaling between 0.85 $\sim$ 1.15, random rotation within 20 degrees
and random translation within 5 voxels.
\subsection{More Implementation Details}
\label{implementation}
As the voxel size varies among different datas, we first preprocessed the training and validation data to the same voxel size of 1mm × 1mm × 1mm. We also adopted a cube bounding
box that is sufficient to hold the annotation mask and cropped the images and ground-truth
masks on both the source domain and target domain.

Our 3D UNet backbone consists of 5 down-sampling blocks and 5 up-sampling blocks
with skip connections. Each down-sampling block of input channel cin and output channel
cout contains one 3D Conv layer of input and output channel cin, one 3D Conv layer of
input channel cin and output channel cout, and two 3D Conv layers of input and output
channel cout. We take batch normalization and ReLU activation after the last three layers
of the network. The up-sampling block is similar to the down-sampling block, except that it
replaces the first 3D Conv layer with the 3D Conv Transpose layer. The number of channels
we take for 3D U-Net is 8, 16, 32, 64, 128, 256. A Softmax layer is applied at the final step.

The structure of the MLM network includes an encoder and a decoder. The encoder has 12 vanilla transformer blocks with embedding dimension of 768, the decoder has 8 vanilla transformer blocks with decoding dimenson of 384. The patch size of MLM is 16 × 16 × 16, and the mask ratio is 75\%.  We use 61 pancreas label masks to train the MLM model about 168,000 iterations. For unseen organ experiments, we only use 1 sample of the target organ to finetune the MLM with about 40 iterations. To be clear, we summarize the overall algorithm on unseen organs in Algorithm \ref{algorithm}.

\subsection{In-domain Pancreas Segmentation}
\label{e1}
\subsubsection{Comparisons with Prior Arts}
\begin{table*}
\centering
\renewcommand\arraystretch{1.4}
\caption{Performance comparison of MLM pipeline with VAE using the different backbones in unseen domain on AMOS. The segmentation results are evaluated with mean Dice score. $\uparrow$ indicates improvement compared to the best comparison method.}
\label{unsenn domain}
\resizebox{\linewidth}{!}{
\begin{tabular}{c|cc|cc|cc} 
\hline
\diagbox{{[}method]}{{[}modality]} & \multicolumn{2}{c|}{~All}                                             & \multicolumn{2}{c|}{Only MRI}                                         & \multicolumn{2}{c}{Only CT}                                            \\ 
\hline
backbone                           & \textbf{3D-Unet}                  & \textbf{Swinunter}                & \textbf{3D-Unet}                  & \textbf{Swinunter}                & \textbf{3D-Unet}                  & \textbf{Swinunter}                 \\
Direct test~~                      & 0.5226                            & 0.4016                            & 0.3456                            & 0.0200                            & 0.6608                            & 0.6995                             \\
VAE  \cite{yao2022unsupervised}                              & 0.6598                            & 0.5487                            & 0.5651                            & 0.3685                            & 0.7377                            & 0.6893                             \\
DISSM   \cite{raju2022deep}                           & 0.6415                            & 0.5225                                 & 0.5516                            & 0.2922                                & 0.7117                            & 0.7022                                 \\
SIFA   \cite{chen2020unsupervised}                             & 0.6005                            & -                                 & 0.5602                            & -                                 & 0.6311                            & -                                  \\
MLM(ours)                          & \textbf{0.6998}($\uparrow$0.0400) & \textbf{0.6014}($\uparrow$0.0527) & \textbf{0.6493}($\uparrow$0.0842) & \textbf{0.4257}($\uparrow$0.0572) & \textbf{0.7393}($\uparrow$0.0015) & \textbf{0.7071}($\uparrow$0.0178)  \\
CT upper bound                     & 0.7167                            & 0.5795                            & 0.5721                            & 0.2759                            & 0.8296                            & 0.8164                             \\
MRI upper bound                    & 0.6913                            & 0.5887                            & 0.8231                            & 0.8372                            & 0.6707                            & 0.5463                             \\
\hline
\end{tabular}
}
\end{table*}

\begin{figure*}[htbp]
\centering
\includegraphics[scale=0.56]{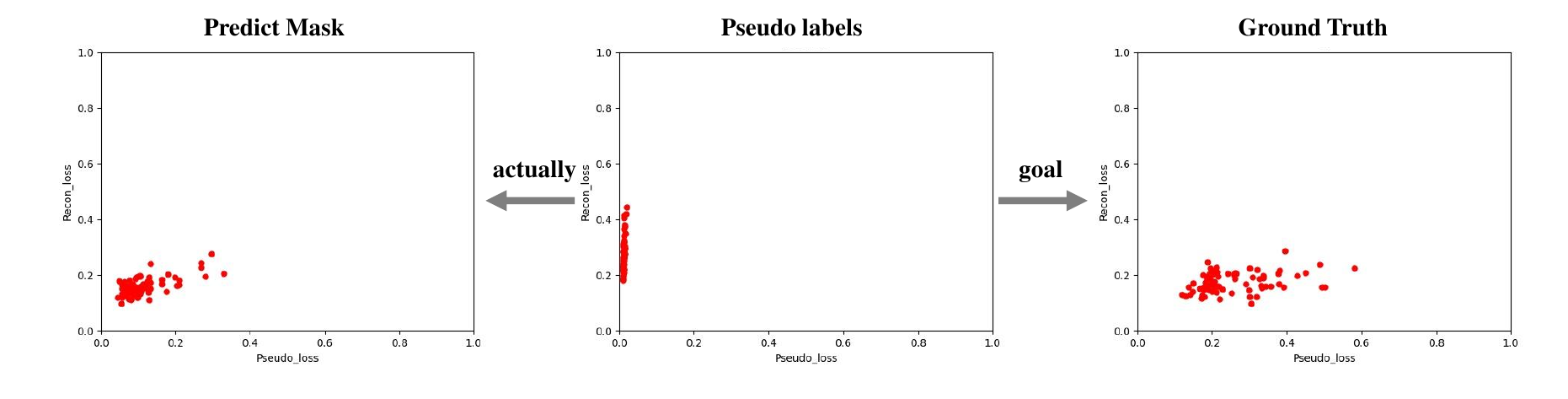}
\caption{Analysis on the distributions of ground truth, pseudo label and
predict mask using data points in MSD validation set. The y-axis represents $\mathcal{L}_{recon}$ and
the x-axis represents $\mathcal{L}_{pseudo}$.}
\label{loss}
\end{figure*}
\begin{table}
\centering
\caption{Ablation study of key components in our proposed MLM pipeline.}
\renewcommand\arraystretch{2}
\label{ablation}
\begin{tabular}{ccc|c}
\multicolumn{1}{l}{\textcircled{1}$\mathcal{L}_{distill}$} & \multicolumn{1}{l}{\textcircled{2}$\mathcal{L}_{pseudo}$} & \textcircled{3}$\mathcal{L}_{recon}$ & \multicolumn{1}{l}{~Dice}  \\ 
\hline
-                                           & -                                          & -                     & 0.6788                     \\
-                                           & -                                          & $\checkmark$          & 0.7355                     \\
-                                           & $\checkmark$                               & -                     & 0.7025                     \\
-                                           & $\checkmark$                               & $\checkmark$          & 0.7466                     \\ 
\hline
$\checkmark$                                 & -                                          & -                     & 0.7355                     \\
$\checkmark$                                & -                                          & $\checkmark$          & 0.7398                     \\
$\checkmark$                                & $\checkmark$                               & -                     & 0.7627                     \\
$\checkmark$                                & $\checkmark$                               & $\checkmark$          & 0.7825                    
\end{tabular}

\end{table}

First, we use the 3D-Unet \cite{cciccek20163d} as the segmentation network backbone, our MLM based on \cite{he2022masked}. And we compare the method which is used the CNN backbone with the same test data. The experiment results are shown in Table \ref{t1}. Among these baseline methods, our propose method is better than the best baseline method in all the target domain datasets. And we use the Swinunter\cite{hatamizadeh2022swin} as the VIT backbone and compare it with the above methods that can replace backbone, except for nnU-Net and SIFA, since both nnU-Net and SIFA have only CNN backbone's and do not have a version with a transformer as a backbon. Table \ref{t1}  illustrates that even with a different backbone our method still achieves excellent results in the target domain, which shows that our proposed MLM pipeline can perform better with the UDA setting than the existing methods.
However, the performance in swinunter backbone is not as good as that in 3D-Unet, which may be due to the fact that medical images segmentation is a data problem rather than a methodology problem, which is well verified in \cite{hofmanninger2020automatic}. The second is that swinunter uses a larger number of parameters requiring more data to train than CNN use. The experiments under VIT backbone also illustrate the good scalability of our MLM.

Figure \ref{result} visualized the segmentation results for each method on the MSD target data, with 3D-Unet as our backbone. Due to the domain gap, both compared methods have a number of false negatives and incomplete shape.  SIFA does not perform well in our experiment, for it only deals with 2D figures which can lead to a lack of overall organ information, which causes a lot of noise and false negatives.
Since both Boundary and Hausdorff distances  constrain  the shape by pixel-wise segmentation output, which can not have the context about the false negative. 
Although both VAE and nnU-Net mitigate this false negative situation to some extent, nnU-Net still has a portion of false negative because nnU-Net's data augmentation does not relate to shape. Although VAE and DISSM introduces the prior information about shape, since VAE and DISSM learns the average organ shape distribution, it also can not obtain the 
context about the false negative. Unlike existing works, we obtain shape using the in-painting approach, which utilizes the self-attention mechanism to capture both short-term and long-term visual dependencies among different patches.  Therefore, MLM is able to complement the existence of false negative patches based on contextual information to guide our segmentation model for better results.



\begin{figure*}
\centering
\includegraphics[scale=0.60]{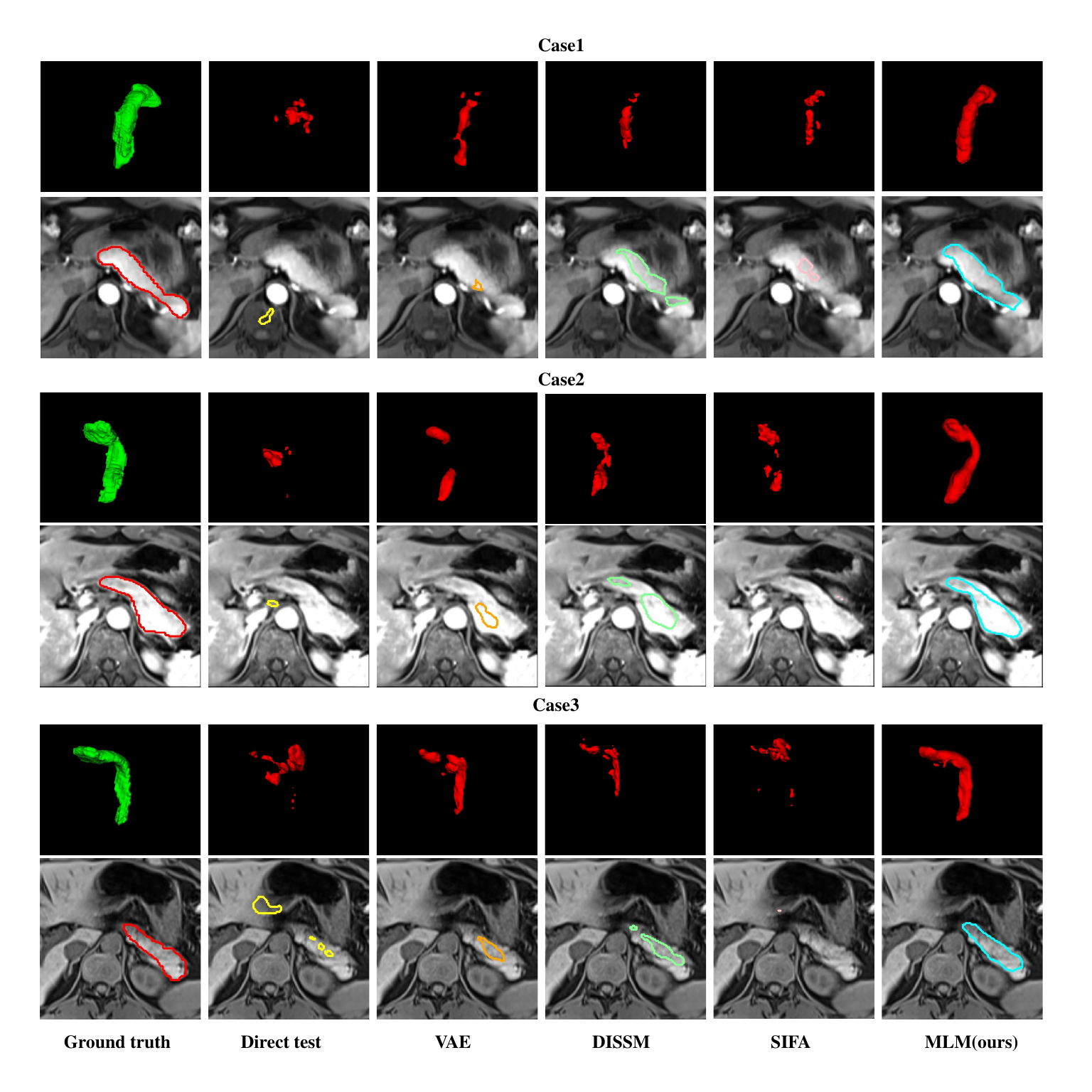} 
\caption{The 3D corresponding 2D visualization for the results on AMOS test cases. For each case, the top image is a 3D visualization, and the bottom image is a corresponding 2D visualization. 
}
\label{aresult1}
\end{figure*}



\subsubsection{Ablation Studies}

We evaluate the components of our network structure, including: \textcircled{1} shape aware self-distillation loss $\mathcal{L}_{distill}$, and \textcircled{2}pseudo loss $\mathcal{L}_{pseudo}$, \textcircled{3}reconstruction loss $\mathcal{L}_{recon}$ in the dual loss optimization of stage 3 on the MSD dataset. Table \ref{ablation} provides a summary. 

To validate the effectiveness of our pipeline, we fine-tune the network on MSD training set with only using $\mathcal{L}_{pseudo}$, $\mathcal{L}_{recon}$ and use two losses without $\mathcal{L}_{distill}$ which respectively yields 0.0567, 0.0237, 0.0678 Dice score improvement compared with the direct test. 
And, applying the $\mathcal{L}_{distill}$ only using $\mathcal{L}_{recon}$ can yield 0.0610 Dice score improvement.
After, we only use $\mathcal{L}_{pseudo}$ after $\mathcal{L}_{distill}$ which can yield the 0.0839 Dice score improvement, the performance is better than only using $\mathcal{L}_{pseudo}$ without $\mathcal{L}_{distill}$. This demonstrates that shape aware self-distillation is an essential step to get a more accurate pseudo model in our pipeline.
In the last row, our MLM pipeline has 0.1037 Dice score improvement which is beneficial to the $\mathcal{L}_{distill}$ and the counterpart relation of $\mathcal{L}_{pseudo}$ and $\mathcal{L}_{recon}$ to avoid the target model find a shortcut solution and punish the segmentation results with low confidence. 

To clearly demonstrate why our method works, we focus on the key losses in our proposed
teacher model. Taking the validation data from MSD dataset as an example, Figure \ref{loss}
illustrates the distribution of data points with regard to their $\mathcal{L}_{recon}$ and $\mathcal{L}_{pseudo}$. The
results of pseudo labels, our predicted masks  and ground truth masks are shown
from left to right. The pseudo labels are initial states for our domain adaptation. Our goal is to
make its distributions of the two losses as similar as that calculated with the ground truth, which is clearly shown in Figure \ref{loss}, i.e., predict mask on the left v.s. ground truth on the right.






\subsection{Generalizing to Unseen Domain}
\label{e2}

In addition to addressing the domain gap caused by the use of different CT devices and protocols in the CT dataset, the domain gap caused by the differences between different modalities is also a problem of concern for us.

Extensive experiments result on AMOS which has both CT and MRI modalities data in pancreas have been shown in Table \ref{unsenn domain}. The methods to constrain the shape by constraining the segmentation boundary, such as boundary and Hausdorff distance, are not applicable to the cross-modal case because the domain gap in the cross-modal is so large that the source model will have so many false negatives that the shape of the segmentation output does not have the original organ shape, and thus the constrained segmentation cannot be improved. Since nnU-Net is not a domain adaptation method and trained on the CT dataset, it is more focused on the CT modality and is not applicable to the data of MRI modality. Thus, we compare the VAE with different backbones on AMOS, and we test the CT upper bound and MRI upper bound  by fine-tuning pretrain source segmentation in the AMOS CT modality and MRI modality respectively. 
\begin{figure*}
\centering
\includegraphics[scale=0.46]{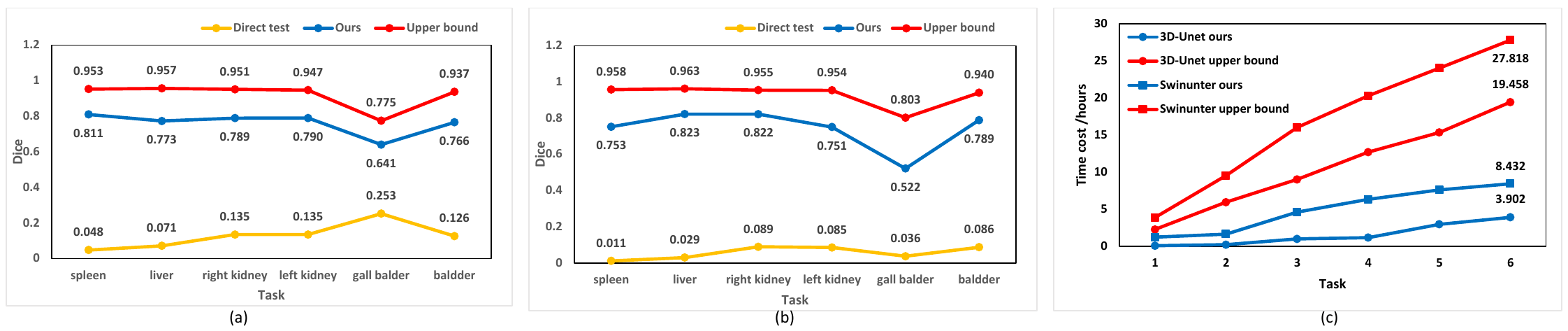} 
\caption{(a) is MLM pipeline performance with 3D-Unet backbone in six unseen organs. (b) is MLM pipeline performance with Swinunter backbone in six unseen organs. (c) is counted the total time spent to train these six models. }
\label{unseen organ}
\end{figure*}

\begin{table*}
\renewcommand\arraystretch{1.4}
\centering
\caption{Performance comparison of MLM pipeline with other UDA segmentation methods using the different dataset as source domain. The segmentation results are evaluated with mean Dice score. $\uparrow$ indicates improvement compared to the best comparison method.}
\label{inter-rater}
\begin{tabular}{c|cc|cc|cc} 
\hline
\multicolumn{1}{l|}{\diagbox{{[}method]}{{[}dataset]}} & \multicolumn{2}{c|}{synapse}                                         & \multicolumn{2}{c|}{WORD}                                                     & \multicolumn{2}{c}{AMOS(only CT)}                                              \\ 
\hline
source dataset                                         & \textbf{NIH}                      & \textbf{MSD}                     & \textbf{NIH}                               & \textbf{MSD}                     & \textbf{NIH}                      & \textbf{MSD}                      \\
Direct test~~                                          & 0.7750                            & 0.7890                           & 0.6809                                     & 0.7614                           & 0.5226                            & 0.6121                            \\
Boundary  \cite{kervadec2019boundary}                                              & 0.7794                            & 0.7929                           & 0.6993                                     & 0.7803                           & 0.6473                                 & 0.7121                                \\
Hausdorff distance \cite{karimi2019reducing}                                     & 0.7863                            & 0.7948                           & 0.7226                                     & 0.7829                           & 0.6998                                & 0.7020                                 \\
VAE  \cite{yao2022unsupervised}                                                   & 0.7867                            & 0.7916                           & 0.7463                                     & 0.7621                           & 0.7377                            & 0.7490                            \\
DISSM  \cite{raju2022deep}                                                & 0.7895                            & 0.7935                           & 0.7480                                     & 0.7790                           & 0.7117                            & 0.7591                            \\
SIFA    \cite{chen2020unsupervised}                                               & 0.7563                            & 0.7780                           & 0.7280                                     & 0.7589                           & 0.6311                            & 0.6817                            \\

nnU-Net \cite{isensee2021nnu}                                              & 0.7706                            & 0.7943                           & 0.7691                                     & \textbf{0.8185(}$\uparrow$0.0160) & 0.6342                               & 0.6239                                 \\
MLM(ours)                                              & \textbf{0.8020}($\uparrow$0.0153) & \textbf{0.8106}($\uparrow$0.0158) & \textbf{\textbf{0.7879}}($\uparrow$0.0188) & 0.8025                           & \textbf{0.7393}($\uparrow$0.0016) & \textbf{0.7667}($\uparrow$0.0076)  \\
upper bound                                            & 0.8037                            & 0.8196                           & 0.8403                                     & 0.8484                           & 0.8296                            & 0.8289                            \\
\hline

\end{tabular}
\end{table*}
Due to the large variety of textures in different modalities, a segmentation model trained in CT modality would have great performance degradation in MRI modality. 
However, the same organ should share the same representation of the 3D anatomy, so we used the shape by VAE and our MLM to guide the segmentation, and the experiment results have been shown in Tabel \ref{unsenn domain}. According to Table \ref{unsenn domain},  The performance of our method is obviously better than sub-optimal VAE in both CT modality and MRI modality. Especially for the  MRI modality our MLM has 0.0842 and 0.0772 Dice score improvement compared with the VAE.
This is due to the fact that we are modeling the shape by in-painting, so our MLM is also able to complete the false negative well when the domain gap is large. 

Figure \ref{aresult1} visualized the segmentation results for each method
on the AMOS target data, with 3D-Unet as our backbone. Due
to the domain gap, both compared methods have a number of
false negatives and incomplete shape. Although the method of SIFA can learn some inter-modal invariance and can reduce the effect of domain gap to a certain extent, the results are not very satisfactory due to the fact that SIFA is processed only for 2D images, which cannot obtain the modal invariance and the spatial location relationship among 3D images. Since VAE and DISSM only learn the global average organ shape which heavily relies on the accurate pseudo labels.
Different with VAE and DISSM, we use MLM for shape completion by in-painting and are able to complement the output of MRI as well as CT based on long-term and short-term contextual information to guide the model for segmentation. This demonstrates the good generalization capability of our MLM pipeline. 




\subsection{Generalizing to Unseen Organ}
\label{e3}


For medical images, in the UDA setting, the main research directions are now focused on unseen datasets or unseen modalities. And according to our best knowledge in UDA setting of 3D medical images, we are the first research to explore unseen organs.

Since the MLM is trained on the pancreas, there is no information about other organs. In order to achieve generalization to other organs, we need to fine-tune the MLM. Specifically, we select a sample from the Synapse dataset and fine-tune MLM to learn organ-specific information. And the pretrain segmentation model still uses the model pretrained on the source domain. After, we validate the fine-tuned MLM on WORD with a specific organ. The detail of the overall algorithm in Algorithm \ref{algorithm}.

According to our best knowledge in the UDA setting of 3D medical images, we are the first research to explore unseen organs, Extensive experiments result on six unseen organs have been shown in Figure \ref{unseen organ}. Unlike the pre-train phase, we only need one unseen organ label mask as support set to finetune MLM. According to Figure \ref{unseen organ} (a) and (b), our MLM pipeline can quickly adapt to new shapes using only one sample of a new organ, which shows that the MLM has a good generalization to unseen organs. As shown in figure \ref{unseen organ} (c), using MLM with either 3D-Unet or Swinunter as the backbone for unsupervised domain adaptation on unseen organs can significantly reduce training time without causing a significant performance drop. Specifically, using 3D-Unet can save nearly six times the time compared to the upper bound while only causing a total performance decrease of 0.6442 across six organs.  Compared to 3D-Unet, Swinunter combines the swin-transformer and CNN, leading to more training time. But our MLM pipeline still can save three times training time. In addition, we also used the fine-tuned MLM model which is fine-tuned by the liver as the MLM to test the performance of the in-pancreas domain on the MSD, Synapse, and WORD dataset with 3D-Unet backbone. The Dice score above these three datasets is 0.7666, 0.8016, and 0.7856. Compare with the Tabel \ref{t1}, after fine-tuning by the liver, MLM still has the pancreas shape information. This reflects that MLM will not have a serious catastrophic forgetting. Links to sample videos of the progressive evolution of the pancreas to other unknown organs can be found in \url{https://drive.google.com/drive/folders/1RwFSIiihTLtBZjmbvGcQxuUSpFnd5iKZ?usp=sharing}.

\subsection{Discussion}

\subsubsection{inter-rater variability}
To verify whether the shape prior is always valid, considering that there is a large inter-rater variability in the labeling of organs by different labelers. We used MSD as our source domain to learn shape information and source domain segmentation model and tested it on Synapse and WORD. Since the pancreas of MSD contains tumor shape, it is very different from normal pancreas to mimic different inter-rater variability, and the results can be seen in Table \ref{inter-rater}. According to Table \ref{inter-rater}, we can learn that whether the normal pancreas shape is used as the source domain or the one with a large difference from the normal shape is used as the source domain to learn the organ shape, the results can be helpful under our setting. In the setting of MSD as the source domain, one of the reasons why nnU-Net on WORD's dataset is superior to our method is that the amount of data in MSD's dataset is much larger than NIH's and the data in MSD's contains tumors and better diversity than NIH's, so when combined with nnU-Net, superior results are achieved.
\subsubsection{limitations and Future work}
Since our work has mainly focused on shape learning of single organs through a novel approach, we have not investigated a priori information on multi-organ shapes and the relative positions of organs, which is a limitation of our work. In future work we would like to explore how to learn the shapes of multiple organs and the corresponding relative positions simultaneously to assist in unsupervised multi-organ segmentation tasks. In addition, we can also combine generative networks to generate different masks to reduce the amount of data required.



\section{Conclusion}
We proposed an unsupervised domain adaptation method to generalize 3D segmentation models to medical images collected from different scanners, protocols, or modalities or organs by in-painting task.
Specifically, we formulated shape modeling as an in-painting task with a new mask strategy in a few label masks. And we further propose a novel shape-aware self-distillation with both in-painting reconstruction loss and pseudo loss, to guarantee its transferability to multiple domains. Our method is inspired by the fact that organs usually show consistent shape, i.e., contours, between most modalities and protocols, while texture and intensity can vary significantly.  Experimental results on three target datasets, different modalities, and different organs demonstrate the superiority and generalization ability of our method. For future work, firstly we can generate synthetic mask data by different generative networks to replace most of the labels, and secondly, we can further investigate the performance of MLM on other organs.
%
%
%
%
\footnotesize
\bibliographystyle{unsrt}

\bibliography{tmi}

\end{document}